\documentclass{article}
\usepackage{ijcai18}
\usepackage{adjustbox}
\usepackage{amsmath}
\usepackage{amssymb}

\usepackage{times}
\usepackage{xcolor}
\usepackage{soul}
\usepackage[utf8]{inputenc}
\usepackage[small]{caption}
\usepackage{makecell}

\title{Scenarios: A New Representation for Complex Scene Understanding}

\author{
Zachary A. Daniels$^1$, 
Dimitris N. Metaxas$^1$
\\ 
$^1$ Department of Computer Science, Rutgers University \\
zad7@cs.rutgers.edu,
dnm@cs.rutgers.edu
}

\begin{document}

\maketitle

\begin{abstract}
The ability for computational agents to reason about the high-level content of real world scene images is important for many applications. Existing attempts at addressing the problem of complex scene understanding lack representational power, efficiency, and the ability to create robust meta-knowledge about scenes. In this paper, we introduce \textbf{scenarios} as a new way of representing scenes. The scenario is a simple, low-dimensional, data-driven representation consisting of sets of frequently co-occurring objects and is useful for a wide range of scene understanding tasks. We learn scenarios from data using a novel matrix factorization method which we integrate into a new neural network architecture, the \textbf{ScenarioNet}. Using ScenarioNet, we can recover semantic information about real world scene images at three levels of granularity: 1) scene categories, 2) scenarios, and 3) objects. Training a single ScenarioNet model enables us to perform scene classification, scenario recognition, multi-object recognition, content-based scene image retrieval, and content-based image comparison. In addition to solving many tasks in a single, unified framework, ScenarioNet is more computationally efficient than other CNNs because it requires significantly fewer parameters while achieving similar performance on benchmark tasks and is more interpretable because it produces explanations when making decisions. We validate the utility of scenarios and ScenarioNet on a diverse set of scene understanding tasks on several benchmark datasets.
\end{abstract}

\section{Introduction}

For many applications (e.g.\ robotics, surveillance, and autonomous vehicles), an agent (either human or machine) must reason about the high-level content of real world scene images. Recently, a lot of progress has been made in constructing algorithms and systems that address fundamental scene understanding tasks such as scene classification, object detection, and semantic segmentation as well as more complex scene understanding tasks such as visual question-answering, automatic relationship extraction, scene graph generation, and learning how to visually reason about objects in simple scenes (e.g. \cite{johnson2016clevr}). While existing methods for solving such tasks are impressive, they can be improved. We propose a new approach for efficiently solving a wide range of tasks related to both simple and complex scene understanding that is complimentary to existing approaches.

\begin{figure*}[t]
\centering
\includegraphics[width=0.8\textwidth]{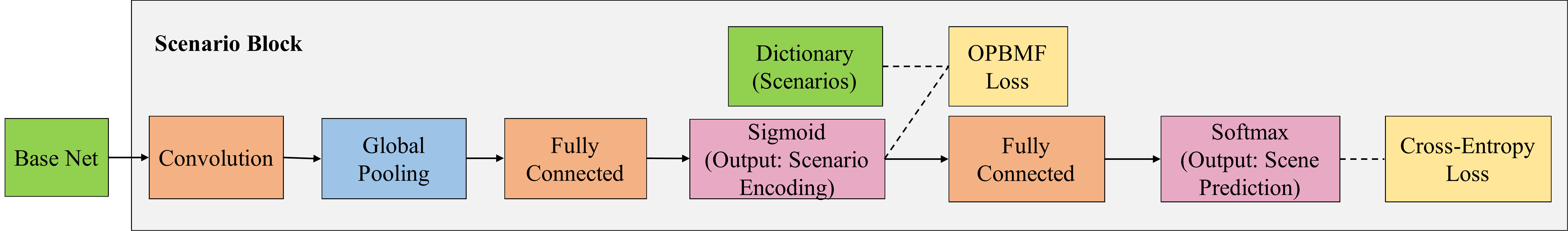}
\caption{The \textbf{scenario block} replaces the final fully connected layers of a standard CNN and consists of: 1) global pooling layers that identify which parts of an image ScenarioNet attends to when recognizing whether a scenario is present in a given image, 2) layers that use a PBMF-based loss function to finetune a dictionary of scenarios and predict the presence of each scenario for a given image, and 3) layers equivalent to multinomial logistic regression that use scenarios as low-dimensional, interpretable features for scene classification.}
\label{fig:scenario_block}
\end{figure*}

We introduce \textbf{scenarios}, a conceptually simple, data-driven representation for complex scene understanding. Scenarios are based on \textit{sets of frequently co-occurring objects}. Scenarios should satisfy a few key properties:
\begin{enumerate}
\itemsep0em 
\item Scenarios are composed of one or more objects.
\item The same object can appear in multiple scenarios, and this should reflect the context in which the object appears, e.g.\ \{keyboard, screen, mouse\} and \{remote control, screen, cable box\} both contain the screen object, but in the first scenario, the screen is a computer monitor, and in the second scenario, it is a television screen.
\item Scenes can be decomposed into combinations of scenarios. For example, a bathroom scene instance might decompose into: \{shower, bathtub, shampoo\} + \{mirror, sink, toothbrush, toothpaste\} + \{toilet, toilet paper\}.
\item Scenarios should be flexible and robust to missing objects. A scenario can be present in a scene without all of its constituent objects being present. 
\end{enumerate}

We propose \textbf{Pseudo-Boolean Matrix Factorization (PBMF)} to identify scenarios from data. PBMF takes a binary \textit{Object-Scene} matrix and decomposes it into 1) a dictionary matrix where each basis vector is a scenario and 2) an encoding matrix that expresses a scene instance as a combination of scenarios. We integrate PBMF into a novel convolutional neural network architecture (CNN), the \textbf{ScenarioNet}. 

ScenarioNet replaces the final convolutional layers in standard CNNs with the \textbf{scenario block} (see Fig.\ \ref{fig:scenario_block}) which consists of three parts: 1) global pooling layers that identify the parts of an image ScenarioNet attends to when recognizing whether each scenario is present in an image, 2) layers that use a PBMF-based loss function to learn a dictionary of scenarios and predict the presence and strength of each scenario for a given image, and 3) layers equivalent to a multinomial logistic regression model that use scenarios as low-dimensional, interpretable features for predicting the scene category. During training, ScenarioNet only requires information about the \textit{presence} of objects in an image and requires no information about the location of objects. For scene classification, class labels are also needed during training. During testing, only images are given.

Using ScenarioNet, we can recover semantic information about scene images at three levels of granularity: 1) scene categories, 2) scenarios, and 3) objects. This allows us to train a single ScenarioNet model capable of performing 1) scene classification, 2) scenario recognition, 3) multi-object recognition, 4) content-based scene image retrieval, and 5) content-based image comparison.

ScenarioNet has several advantages over other CNNs. It is more computationally efficient than other CNNs because it requires significantly fewer parameters in order to achieve similar performance on benchmark tasks, and it is more interpretable because it produces explanations when making decisions. For example, for scene classification, predicted scenarios are used as low-dimensional, interpretable features; humans can verify the presence of each predicted scenario in an image by examining the scenario-localizing attention maps produced by the network; and humans can inspect how much influence each scenario exerts when assigning a class. This allows us to understand how a network arrives at specific decisions and debug it when it makes incorrect decisions.

We evaluate the utility of ScenarioNet using the SUNRGBD \cite{song2015sun}, ADE20K \cite{zhou2017scene}, and MIT 67 Indoor Scenes \cite{quattoni2009recognizing} datasets. We perform quantitative experiments on multi-object recognition, scene classification, and content-based image retrieval, and show that ScenarioNet performs comparable to and often better than existing methods. We also show examples demonstrating the interpretability and expressiveness of ScenarioNet.

\section{Related Work}

Discovering meaningful groups of objects is not a new idea. The simplest object-based representations are those that utilize pairwise co-occurrence relationships between objects (e.g. \cite{rabinovich2007objects}). Scenarios go one step further by efficiently learning groups of objects of varying size. Many works focus on hierarchical models relating objects and scenes. \cite{fengsemantic} constructs a tree-based hierarchy of concepts based on object co-occurrence graphs. Objects sharing an ancestor node can be grouped into scene concepts, an idea similar to our scenarios. Several issues exist with using a tree structure for specifying scene concepts. To compute explicit scenarios, one must identify where to cut the tree. Additionally, while individual concepts can belong to multiple scene concepts by cutting the tree at different ancestor nodes, it becomes hard to properly place objects in the hierarchy that serve different functions within different groups, e.g.\ a screen with a keyboard and mouse is different from a screen with a cable box and remote. Our scenarios address these issues and provide additional information, e.g. how important each object is to a given scenario and how to decompose scene instances into combinations of scenarios. Other tree-based and hierarchical models for scene understanding exist. \cite{choi2012tree} introduces a tree structure where nodes represent objects and latent variables and edges represent positive and negative correlations between nodes. These trees implicitly capture scenarios while our work learns explicit scenarios. \cite{fan2008integrating} exploit hierarchies of concepts to build ontologies for content-based image retrieval. \cite{lan2013subcategories} investigate context at three levels: individual objects, parts of objects, and visual composites.

Other groups focus on using sets of objects to aid object detection. \cite{li2012automatic} discovers groups of objects of arbitrary size, model these groups using deformable parts models, and directly detects these groups in images. \cite{cinbis2012contextual} constructs classifiers that operate over sets of objects using object-object and object-scene relations to re-score and remove noisy detections.

\section{Proposed Method}
\subsection{Identifying Scenarios from Data: Pseudo-Boolean Matrix Factorization}

We now discuss the technical details of our model. We begin by focusing on the question: how do we identify which sets of objects naturally group together to form scenarios? We start with a training set of scene instances and a finite set of predetermined objects. We have ground-truth annotations for the presence (or lack thereof) of every object in every scene instance given by either humans or object detectors. For each training instance, we create a vector of object presences where each element corresponds to a specific object, and the element is 1 if the object is present and 0 otherwise. We concatenate these vectors to form a matrix $A$ where each row corresponds to a specific object and each column is a training instance. After specifying the number of desired scenarios $k$ (which can be estimated from the data), we decompose $A$ into two smaller approximately binary matrices: a dictionary matrix $W$ representing a set of scenarios and an encoding matrix $H$ that expresses scene instances as combinations of scenarios. Each column of $W$ represents a single scenario and each row represents an object. If element $W_{ij}$ is 0 or very small, object $i$ is not present in scenario $j$. The closer $W_{ij}$ is to 1, the more important object $i$ is to scenario $j$. Each column of $H$ represents a specific scene instance and each row represents a specific scenario. If element $H_{ij}$ is 0 or very small, then scenario $i$ is not present in scene instance $j$. The closer $H_{ij}$ is to 1, the more important scenario $i$ is to scene instance $j$. We now face the question of which decomposition to use.

\subsubsection{Formulation of PBMF}

We propose identifying scenarios using an approximation of Boolean matrix factorization (BMF) \cite{miettinen2008discrete}. In BMF, $A$, $W$, and $H$ are binary matrices and the matrix multiplication is Boolean (denoted as $\circ$):
\begin{align}
\scriptsize
\begin{split}
\scriptsize
&\min_{W,H} ||(A-W\circ H)||_{1} \; s.t. \; W \in \{0,1\}, H \in \{0,1\}
\end{split}
\label{eq:BMF}
\end{align}
BMF is well-suited for identifying scenarios from data because: 1) it efficiently compresses and preserves information using low-dimensional representations; 2) the basis vectors are easy to interpret; 3) it discovers meaningful interactions between objects; and 4) the encoding vectors are sparse, so each instance is expressed by a small subset of scenarios.

We use a gradient descent-based approach to solve the optimization problem. The optimization problem in Eq. \ref{eq:BMF} is not continuous, so we approximate Boolean matrix multiplication as $W\circ H \approx \min(WH,1)$ and relax the constraints to lie in $[0,1]$. However, using $\min(WH,1)$ results in cases where the gradient dies, so we further approximate $\min(WH,1) \approx \min(WH,1 + 0.01WH)$. Our basic \textbf{Pseudo-Boolean Matrix Factorization (PBMF)} formulation becomes:
\begin{align}
\scriptsize
\begin{split}
\scriptsize
&\min_{W,H} ||(A-\min(WH,1+0.01WH))||_{F}^{2} \; s.t. \; W \in [0,1], H \in [0,1]
\end{split}
\label{eq:PBMF}
\end{align}

The above formulation (Eq. \ref{eq:PBMF}) is still not perfectly suited for discovering scenarios. We add three additional terms: an orthogonality penalty to encourage diversity between scenarios and sparse penalties on the scenario dictionary and encodings to make $W$ and $H$ closer to binary matrices and improve interpretability. We also introduce $\Omega$ which is a weight matrix that decreases the importance of very common objects and increases the importance of rare objects.
\begin{align}
\scriptsize
\begin{split}
\scriptsize
&\min_{W,H} ||\Omega \; \bullet \; (A-\min(WH,1+0.01WH))||_{F}^{2} \\
& + \alpha_{1}||W^{\intercal}W - diag(W^{\intercal}W)||_{F}^{2} + \alpha_{2}||W||_{1} + \alpha_{3}||H||_{1} \\
& s.t. \; W \in [0,1], H \in [0,1], \\
&\Omega_{ij} = \max\left(A_{ij} * \left(1 + \log\left(\frac{N_{instances}}{N_{objects}}\right)\right),1\right)
\end{split}
\label{eq:PBMF_Full}
\end{align}
$\bullet$ denotes element-wise matrix multiplication. The $\alpha$s represent tradeoff parameters.

\subsection{ScenarioNet: Updating and Recognizing Scenarios from Visual Data}

So far we've assumed we have perfect knowledge of all ground-truth object data. This means that if we're given a previously unseen scene instance, we can hold the scenario matrix constant and directly solve for the encoding matrix. In practice, we'll not have object data at test time. We need to learn how to recover the scenario encoding for a specific scene instance entirely from visual data. To do this, we integrate PBMF with CNNs. We propose \textbf{ScenarioNet}, a convolutional neural network that learns to identify and recognize scenarios from real-world visual data, performs scene classification using the predicted scenario encoding, and generates attention maps that explain why the net thinks a specific scenario is present in a given image. \textbf{ScenarioNet learns to predict the scenario encoding matrix $\hat{H}$ and finetunes the dictionary $W$ so it adapts to the noisier $\hat{H}$. $W$ also incorporates feedback from the scene classification task to improve discriminability}. The key architectural difference between ScenarioNet and other CNNs is the \textbf{scenario block} (see Fig.\ \ref{fig:scenario_block}) which replaces the final fully connected layers used for classification in standard CNNs.

We now describe the rationale behind the scenario block. The final convolutional layers of a neural net such VGGNet are fed into a global average pooling layer. This layer in combination with the class activation mapping technique \cite{zhou2016learning} allows us to identify which parts of an image ScenarioNet attends to when determining if a scenario is present in the image. The output of the global pooling layer is fed into a fully connected layer followed by a sigmoid transformation layer. The sigmoid layer outputs the \textbf{scenario encoding vector} and enforces each element of the vector is between 0 and 1. This vector tells us how present each scenario is in a given image. The scenario encoding layer feeds into a PBMF loss layer which finetunes the scenario dictionary and provides feedback to the network. The scenario encoding is also fed into a sequence of layers equivalent to a multinomial logistic regression model that uses scenarios as low-dimensional, interpretable features for scene classification.

\subsubsection{Training ScenarioNet}
Training ScenarioNet is slightly more complicated then training a standard CNN. During training, ScenarioNet only requires information about the \textit{presence} of objects in an image and requires no information about the location of objects. For scene classification, class labels are also needed during training. During testing, only images are given. The first step of training ScenarioNet involves learning the scenario dictionary using ground-truth object presence data. Then, we train the net to predict the scenario encodings while finetuning the dictionary. Next, we train a softmax classifier for scene classification on top of a frozen net. Finally, we jointly finetune the net for scenario recognition and scene classification while once again finetuning the dictionary. It is useful to finetune only the last few convolutional layers of networks that have been previously trained for scene classification (e.g.\ on the Places dataset \cite{zhou2017places}) since scenario recognition and scene classification are closely related. Typically, each step of the finetuning process takes between 10 and 20 iterations. To finetune the dictionary while training the net, we use alternating projected gradient descent. During training, we hold the scenario dictionary constant and finetune the network using backpropagation in mini-batches to predict the encoding coefficients. After every four iterations, we hold the network constant and perform a full pass through the data to reconstruct $\hat{H}$ and finetune the scenario dictionary $W$ using projected gradient descent. Alternatively, $W$ can be efficiently finetuned using mini-batches by noting that the gradient of the PBMF loss w.r.t. $W$ is able to be decomposed as a sum of gradients over sub-batches of $\hat{H}$; thus, we never have to compute the full $\hat{H}$ at any point in time.

\subsubsection{Interpreting the Output of ScenarioNet}

\begin{figure}[ht!]
\small
\centering
\includegraphics[width=0.47\textwidth]{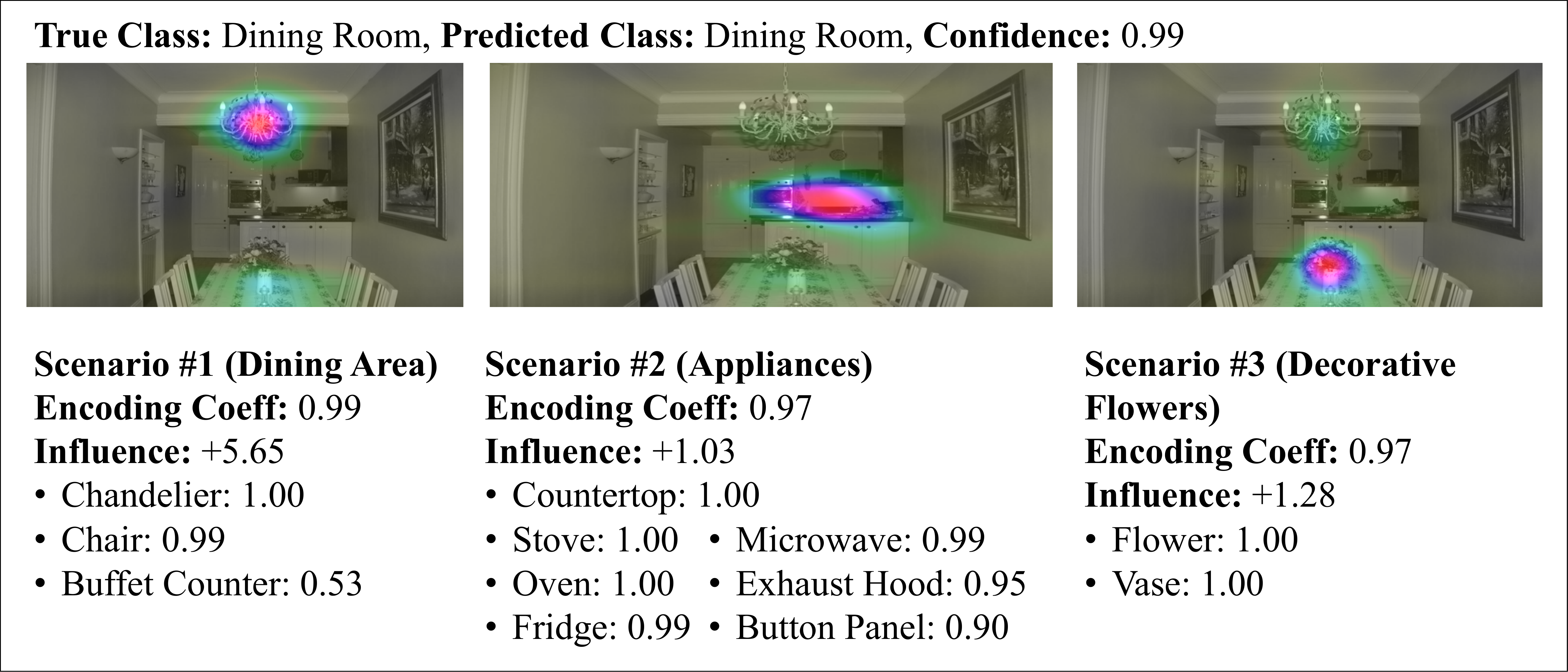}
\caption{We demonstrate the interpretability of ScenarioNet. We show the top-3 predicted scenarios for a dining room scene along with the corresponding activation maps. Please view in color.}
\label{fig:explanation}
\end{figure}

Given an input image, ScenarioNet provides us with a probabilistic scene class assignment, a vector of scenario encoding coefficients, the dictionary of scenarios, and activation maps that can be used to localize the discriminative parts of each scenario. In Fig.\ \ref{fig:explanation}, we show an example of decomposing a scene instance into its top-3 strongest detected scenarios using ScenarioNet. We see that ScenarioNet correctly predicts with high confidence that the scene category is ``dining room''. The top-3 scenarios support this: one focuses on dining areas, one on kitchen appliances, and one on decorative flowers. The encoding coefficient denotes the strength of each scenario. Note that all of the encoding coefficients are close to one since these are the strongest detected scenarios. As this coefficient decreases, the scenarios become less present. Encoding coefficients tend to cluster around 0 and 1. Recall that ScenarioNet uses scenarios as features for scene classification. We can define a scenario's \textit{influence score} for a specific class to be the corresponding weight in the multinomial logistic regression model. If the influence is a large positive number, the scenario provides strong evidence for the specified class. If it is a large negative number, the scenario is strong evidence against a specific class. For this image, scenario 1 is very indicative of the scene class, while scenarios 2 and 3 are weakly indicative. We can also see how important each object is to each scenario. For example, in scenario 1, the ``chandelier'' and ``chair'' objects are more important to defining the scenario than the ``buffet counter'' object. By examining the scenario activation maps, we see that each predicted scenario is present and net attends to regions of the image containing objects present in the scenarios.

As machine learning models become more complex, greater importance should be placed on learning explainable models and making complex models more interpretable. This is especially important for visual recognition tasks where deep neural networks are popular. As the previous example demonstrates, ScenarioNet is highly interpretable. This is a key advantage over other CNNs.

\subsubsection{Efficiency}
It is also interesting to note that our network has significantly fewer parameters than equivalent base architectures. For example, the final convolutional layers of VGG-16 typically consist of a 4096-by-4096 matrix followed by a 4096-by-\#classes matrix for a total of $4096(4096+\#classes)$ parameters. Our net uses a 512-by-\#scenarios matrix followed by a \#scenarios-by-\#classes matrix for a total of $\#scenarios(512 + \#classes)$ parameters. Since \#scenarios $<<$ 4096 (we use between $k=25$ and $k=70$ scenarios in our experiments), this results in over a 100x reduction in the number of parameters in the final layers. This reduces the memory footprint of the \textit{total} net by a factor of \~{}10, and the net is \~{}15\% faster during testing. 

\section{Experimental Results and Analysis}

In this section, we evaluate the reconstruction ability of PBMF, and also analyze the performance of ScenarioNet on three common scene understanding tasks: multi-object recognition, scene classification, and content-based scene image retrieval. We first explain the general experimental setup.

\subsection{Experimental Setup}
\label{experimental_setup}
We conduct experiments on the SUNRGBD \cite{song2015sun}, ADE20K \cite{zhou2017scene}, and MIT 67 Indoor Scenes \cite{quattoni2009recognizing} datasets. We divide each dataset into separate training and test sets using the recommended splits for the SUNRGBD and MIT67 datasets and a random split for the ADE20K dataset. For each dataset, we only consider objects that appear in at least 1\% of the training instances resulting in 55 objects for SUNRGBD, 193 for ADE20K, and 166 for MIT67. We use random cropping and horizontal mirroring to augment the training examples. For the SUNRGBD dataset, we use the 15 most frequently occurring scene classes, reserving 100 samples per class for test data, and generating 1000 samples per class for the training data. For the ADE20K dataset, we use the 31 most frequently occurring scene classes, reserving 25 samples per class for test data, and generating 500 samples per class for training data. For the MIT67 dataset, we use 67 scene classes, reserving 20 samples per class for test data, and generating 800 samples per class for training data. \textbf{We learn 25 scenarios for SUNRGBD, 70 for ADE20K, and 70 for MIT67. We use VGG-16 as our base CNN architecture}, replacing the final fully-connected layers with the scenario block. \textbf{For the MIT dataset, we only have object annotation data for about one-fifth of the training data, the amount of annotated data is very imbalanced between classes, and the annotations are much noisier than for the other datasets. These properties make learning scenarios on the MIT dataset much more difficult than for the other datasets, but we are still able to achieve relatively good results.} For this dataset, we learn the scenarios using the annotated portion of the dataset and train a scene classifier on top of these scenarios for the full dataset. 

\subsection{Reconstruction Error of PBMF}
\begin{figure}[ht!]
\centering
\includegraphics[width=0.33\textwidth]{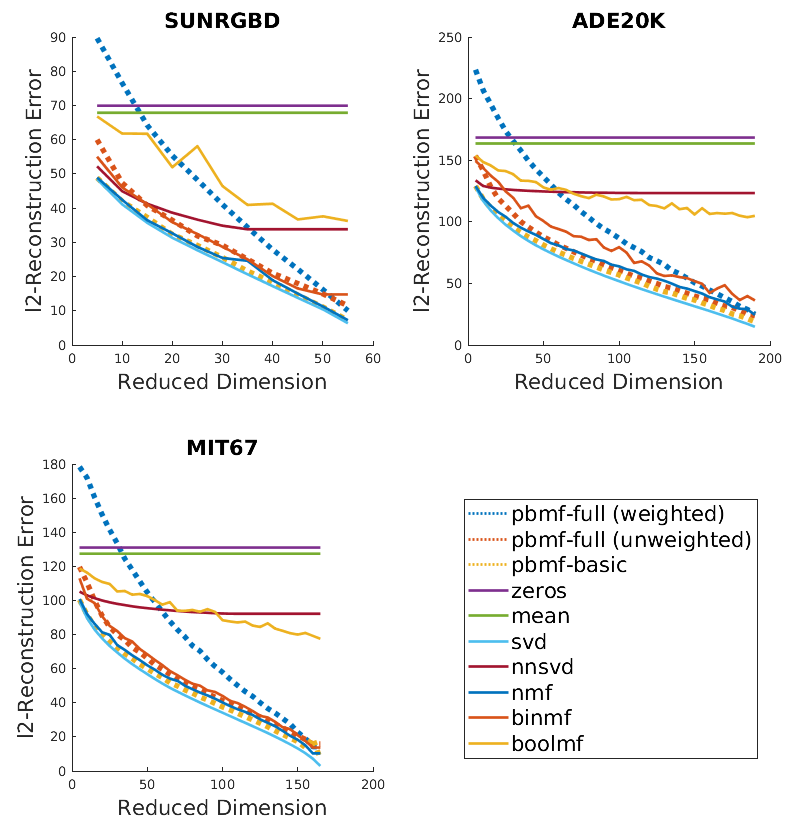}
\caption{Reconstruction error between a recovered and groundtruth matrix as the dimensionality of the reduced representation is varied}
\label{fig:recon}
\end{figure}

PBMF is a lossy factorization. We want to determine how much information about object presence is lost as a result of the decomposition. \textbf{For this experiment, we assume perfect, ground-truth knowledge of the object presences.} We consider three cases of PBMF: PBMF-Basic (Eq. \ref{eq:PBMF}) and PBMF-Full (Eq. \ref{eq:PBMF_Full}) with uniform weighting and PBMF-Full using the proposed weight matrix. We compare to the SVD, NNSVD \cite{ding2006orthogonal}, NMF \cite{paatero1994positive}, Greedy Boolean MF \cite{miettinen2008discrete}, and Binary MF \cite{zhang2007binary} as well as all-zeros and all-mean values baselines. We initialize the basis and encoding matrices using a procedure similar to \cite{zhang2007binary}. Results are plotted in Fig.\ \ref{fig:recon}. PBMF-Basic works exceptionally well for reconstruction, but as we add additional constraints (particularly when we increase the weights of rare objects), the reconstruction worsens. However, the reconstruction error remains tolerable, especially for reasonable numbers of scenarios, so we feel PBMF is suitable for learning scenarios. 

\subsection{Multi-Object Recognition from Scene Images}
\label{exp:mor}
In addition to noise from the lossy PBMF, we also need to consider noise resulting from mapping visual data to objects. We consider the task of tagging images with their constituent objects. We use ScenarioNet to predict the scenario encoding matrix $\hat{H}$ and try to recover the object matrix $A \approx W\hat{H}$. We compare against a finetuned object detector \cite{redmon2016yolo9000} and VGG-16 finetuned for multi-object recognition. We use the macro (averaged) area under the precision-recall curve as our metric since the data is very imbalanced for rare objects. In Table \ref{tab:objects}, we show results for when we consider objects that appear in at least 1\% of data (very rare) and 5\% of data (rare) to show how imbalance affects both methods. ADE20K tends to contain smaller objects and parts-of-objects, so the VGG-based nets outperform the object detector on this data, and perform worse on SUNRGBD. VGG-Objects and ScenarioNet perform similarly despite PBMF being lossy and the output of ScenarioNet being 2-3 times smaller in dimensionality. Interestingly, ScenarioNet performs better than VGG-Objects on the SUNRGBD-5\% task. We believe ScenarioNet performs well because it excels at capturing context and because it is easier to recognize scenarios (defined by a few key objects) than individual objects. ScenarioNet has several advantages over individual object-based methods: it finds meaningful relationships between objects and captures global scene information.

\begin{table}[h!]
\centering
\begin{adjustbox}{width=0.46\textwidth}
\begin{tabular}{|l||r|r||r|r|}
\hline
 & \multicolumn{2}{|c|}{SUNRGBD} & \multicolumn{2}{|c|}{ADE20K} \\ \hline \hline
Method & \makecell{1\% \\ (55 Objs)} & \makecell{5\% \\ (16 Objs)} & \makecell{1\% \\ (193 Objs)} & \makecell{5\% \\ (50 Objs)} \\ \hline \hline
Random & 0.066 & 0.152 & 0.070 & 0.171 \\ \hline
Object Detection (YOLOv2) & 0.442 & 0.633 & 0.379 & 0.587 \\ \hline
VGG-Objects & 0.369 & 0.574 & 0.475 & 0.696 \\ \hline \hline
\textbf{ScenarioNet} & \textbf{0.356} & \textbf{0.585} & \textbf{0.452} & \textbf{0.683}\\ \hline
\end{tabular}
\end{adjustbox}
\caption{Macro-AUPRC for multi-object recognition}
\label{tab:objects}
\end{table}

\subsection{Scene Classification}
We now consider the task of scene classification where we care more about global scene information than local objects. In the following sections, we compare ScenarioNet to other object-based representations, baseline CNNs, compressed CNNs, and other mid-level features. Results are reported in Table \ref{tab:scenes}. For experiments not involving a CNN, we train a logistic regression model on top of the given features.

\begin{table}[h!]
\centering
\begin{adjustbox}{width=0.46\textwidth}
\begin{tabular}{|l||r||r|r|r|}
\hline
\multicolumn{5}{|c|}{\textbf{\textit{Object-Based Representations}}} \\ \hline
Method & Dimens. & SUNRGBD & ADE20K & MIT \\ \hline \hline
Object Bank + PCA & 8000 & 0.296 & 0.511 & 0.39\\ \hline
Object Detection (YOLOv2) & 55/193/166 & 0.399 & 0.639 & 0.517\\ \hline
VGG-Objects & 55/193/166 & 0.483 & 0.726 & 0.6187\\ \hline \hline
\multicolumn{5}{|c|}{\textbf{\textit{Baseline CNNs}}} \\ \hline
AlexNet & 4096 & 0.469 & 0.786 & 0.687\\ \hline
GoogLeNet & 2048 & 0.541 & 0.796 & 0.737\\ \hline
VGG-16 & 4096 & 0.531 & 0.809 & 0.792\\ \hline
ResNet-50 & 1024 & 0.509 & 0.777 & 0.687\\ \hline\hline
\multicolumn{5}{|c|}{\textbf{\textit{Dimensionality-Reducing and Lower-Parameter CNNs}}} \\ \hline
VGG-Reduced & 25/70/70 & 0.458 & 0.787 & 0.722 \\ \hline
VGG-GAP & 512 & 0.486 & 0.767 & 0.779\\ \hline
VGG-GMP & 512 & 0.463 & 0.786 & 0.723\\ \hline \hline
\multicolumn{5}{|c|}{\textbf{\textit{Attribute-Based Representations}}} \\ \hline
SUN-Attribute & 102 & 0.429 & 0.705 & 0.655\\ \hline
Classemes & 2659 & 0.309 & 0.581 & 0.448\\ \hline
Meta-Classes & 15232 & 0.36 & 0.635 & 0.525\\ \hline \hline
\multicolumn{5}{|c|}{\textbf{\textit{Learned Mid-Level Visual Representations}}} \\ \hline
Mid-Level Patches & 14070 & N/A & N/A & 0.381* \\ \hline
Mid-Level Vis. Elem. & 67000  & N/A & N/A & 0.64* \\ \hline
DPM & N/A & N/A & N/A & 0.304* \\ \hline
RBoW & N/A & N/A & N/A & 0.379* \\ \hline
BoP & 3350 & N/A & N/A & 0.461* \\ \hline
Discriminative Parts & 4926 & N/A & N/A & 0.514* \\ \hline \hline
\multicolumn{5}{|c|}{\textbf{\textit{Proposed Model}}} \\ \hline
\textbf{ScenarioNet} & \textbf{25/70/70} & \textbf{0.520} & \textbf{0.794} & \textbf{0.725}\\ \hline
\end{tabular}
\end{adjustbox}
\caption{Scene classification accuracy; * denotes reported results}
\label{tab:scenes}
\end{table}

\subsubsection{Comparison to Other Object-Based Representations}
We first consider object-based representations. These include the same models as in Sec.\ \ref{exp:mor} (using the object probabilities as features) and also Object Bank features \cite{li2010object} compressed to 8000 dimensions using PCA. ScenarioNet is significantly better than all other object-based representations for scene classification despite its lower dimensionality. This suggests that scenarios are better at capturing global scene information than individual object-based approaches. This is partly because ScenarioNet is trained to jointly recognize objects and scenes, a key difference to the other methods.

\subsubsection{Comparison to Baseline CNNs}

CNNs are currently the most popular method for scene classification. We finetune AlexNet \cite{krizhevsky2012imagenet}, GoogLeNet \cite{szegedy2015going}, and VGG-16 \cite{simonyan2014very} models that have been pre-trained on the Places dataset \cite{zhou2017places} as well as a ResNet-50 CNN \cite{he2016deep} pre-trained on ImageNet \cite{deng2009imagenet}. Since ScenarioNet extends VGG-16, we focus on how these two nets compare. ScenarioNet tends to slightly underperform VGG-16 by about 1-2\%. The most significant drop in performance is on the MIT dataset, but recall that ScenarioNet is forced to learn scenarios on a significantly smaller, noisier, and imbalanced subset of the training data (see Sec.\ \ref{experimental_setup}). ScenarioNet has several major advantages over VGG-16; it has significantly fewer parameters, has the ability to explain its decisions, and can produce scenario encodings which are useful for tasks beyond scene classification. In the next section, we will see that ScenarioNet generally performs better than standard VGG-16 nets that compress the feature space to the same dimensionality as ScenarioNet.

\subsubsection{Comparison to Dim-Reducing and Lower-Parameter CNNs}

We modify VGG-16 so the output of the final feature layer is the same dimensionality as our scenario-based representation by shrinking the final fully-connected layers. ScenarioNet matches or outperforms the compressed VGG-16 net in all cases, including on MIT67 where the VGG-Reduced net's representation is learned on significantly more and better balanced data. This might be because ScenarioNet constrains the intermediate representation to have high-level meaning while the compressed-VGG net lacks such guidance, making it susceptible to finding worse local minimum. We also compare ScenarioNet to VGG nets which replace the double fully-connected layers with global average pooling (GAP) and global max pool (GMP) layers. These nets contain roughly the same number of parameters as ScenarioNet. In five of six cases, we outperform or match the low-parameter nets.

\subsubsection{Comparison to Mid-Level Representations}
Finally, we compare against three other types of mid-level representations: attributes, mid-level visual patches, and parts-based models. Attributes are high-level semantic properties shared between multiple classes \cite{farhadi2009describing}. We consider three attribute-like representations: SUN Attributes \cite{patterson2012sun}, Classemes \cite{torresani2010efficient}, and Meta-Classes \cite{bergamo2012meta}. Several representations consider visually-distinct, meaningful mid-level patches \cite{singh2012unsupervised} and mid-level visual elements \cite{doersch2013mid}. Finally, we consider parts-based models including the deformable parts model (DPM) \cite{pandey2011scene}, reconfigurable bags-of-words (RBoW) \cite{parizi2012reconfigurable}, bags-of-parts (BoP) \cite{juneja2013blocks}, and discriminative parts \cite{sun2013learning}. We outperform all of these methods, but it should be noted that for the non-attribute-based features, we use the reported results on the MIT dataset because the code to generate these features is either unavailable or prohibitively expensive to run on our machines. It should also be noted that these methods pre-date CNNs, and not all of the reported results include the use of training data augmentation while ScenarioNet does.

\subsection{Content-Based Querying and Comparison}
ScenarioNet is useful for content-based scene image retrieval because it can retrieve images satisfying a set of high-level criteria based on the scene category, scenarios, and objects present in an image (e.g. find images of scene category A OR B THAT CONTAIN scenarios X AND Y but EXCLUDE object Z). Often, we want to query for broad concepts and not individual objects. Scenarios offer a nice compromise between global (scene category) and local (object) information. It is easy for humans to examine the scenario dictionary and form complex queries because scenarios are low-dimensional and interpretable. Scenarios can also act as an efficient hashing mechanism because they are low-dimensional and approximately binary, so memory requirements are low and retrieval can be performed in an efficient manner.

In Table \ref{tab:complex_querying}, we evaluate ScenarioNet for complex content-based scene image retrieval. We form 500 random queries, each consisting of a desired scene class, two objects that should be present, and one object that should be absent but frequently co-occurs with the other two objects, i.e. $(SC \cap O_1 \cap O_2 \cap \neg O_3)$. We do not consider querying against scenarios for this task because no other method besides ScenarioNet is capable of recognizing scenarios. We measure the relevance of a returned image as the proportion of query terms that are satisifed. We compute the normalized discounted cumulative gain for the top-5 result images for each query. ScenarioNet is very competive with the other methods, matching VGG-Objects for the best performance on ADE20K, and coming very close to both baselines on SUNRGBD.

\begin{table}[h!]
\centering
\begin{adjustbox}{width=0.36\textwidth}
\begin{tabular}{|l||r|r|}
\hline
Method & SUNRGBD & ADE20K \\ \hline \hline
Random & 0.302 & 0.313 \\ \hline
Object Detection (YOLOv2) & 0.679 & 0.760 \\ \hline
VGG-Objects & 0.686 & 0.799\\ \hline \hline
\textbf{ScenarioNet} & \textbf{0.652} & \textbf{0.799}\\ \hline
\end{tabular}
\end{adjustbox}
\caption{NDCG@5 for retrieving images of a given class containing 2 specific objects and not containing a third highly-correlated object}
\label{tab:complex_querying}
\end{table}

\begin{figure}[ht!]
\centering
\includegraphics[width=0.4\textwidth]{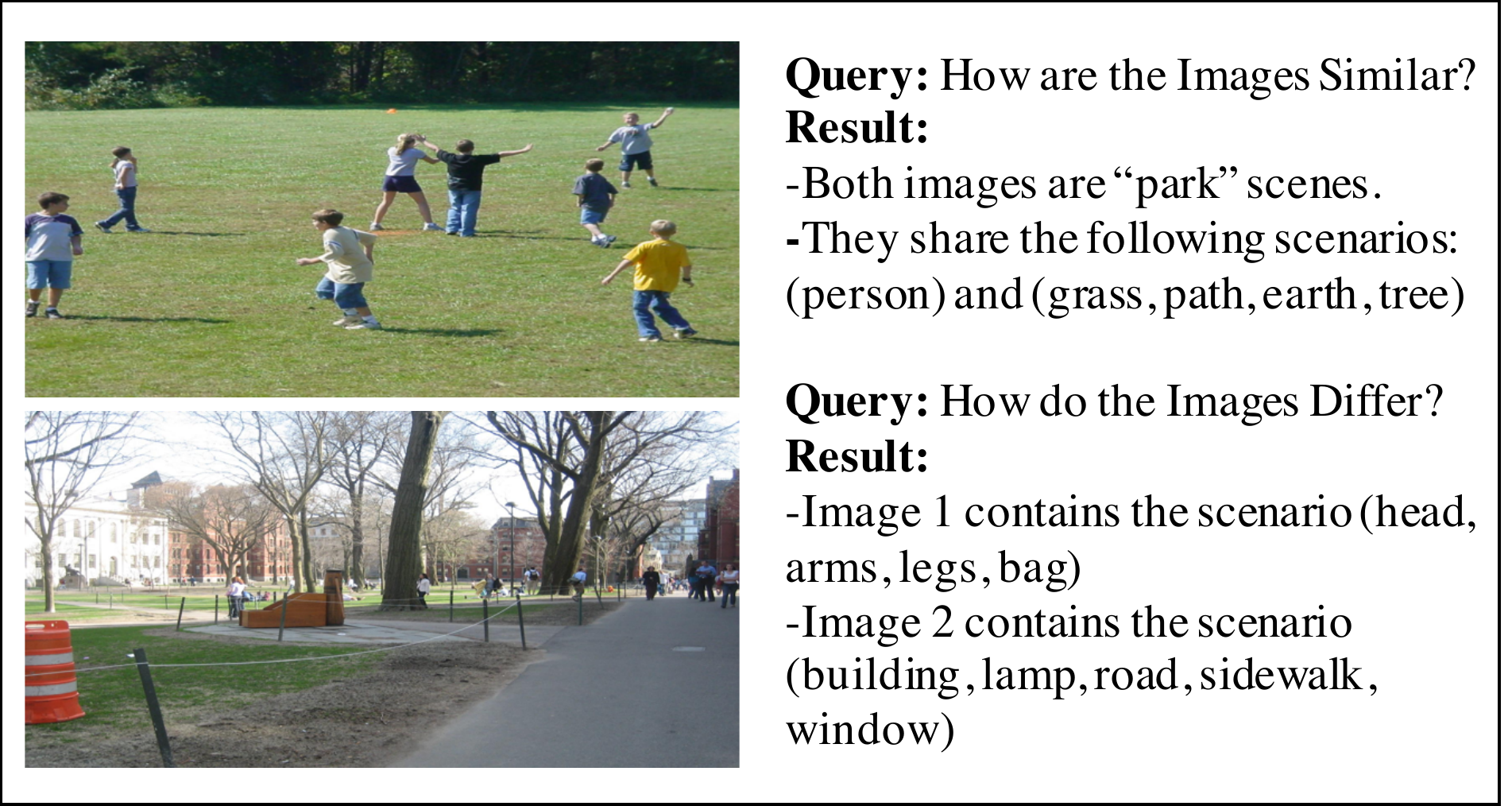}
\caption{Using ScenarioNet to find high-level similarities and differences between two images.}
\label{fig:complex_2}
\end{figure}

ScenarioNet is also useful for generating a quick overview of the similarities and differences between two scene images without relying on (often unnecessary) information about individual objects. Fig.\ \ref{fig:complex_2} shows an example.

\section{Conclusions}

We introduced scenarios as a new way of representing scenes. The scenario is a simple data-driven representation based on sets of frequently co-occurring objects. We provided a method for learning scenarios from data by combining PBMF with CNNs to form the ScenarioNet. Our experiments showed that a single ScenarioNet model can perform scene classification, scenario recognition, multi-object recognition, content-based scene image retrieval, and content-based image comparison with performance comparable to or better than existing models. We showed that scenarios have several advantages over individual object-based representations; specifically, they are lower-dimensional, capture global scene context, and find relationships between objects. We also discussed and demonstrated the computational efficiency and interpretability of ScenarioNet compared to traditional CNNs.


\clearpage
\newpage
{\fontsize{10}{10}\selectfont
\bibliographystyle{named}
\bibliography{scenarios.bbl}

\begin{thebibliography}{}

\bibitem[\protect\citeauthoryear{Bergamo and Torresani}{2012}]{bergamo2012meta}
Alessandro Bergamo and Lorenzo Torresani.
\newblock Meta-class features for large-scale object categorization on a
  budget.
\newblock In {\em CVPR}, 2012.

\bibitem[\protect\citeauthoryear{Choi \bgroup \em et al.\egroup
  }{2012}]{choi2012tree}
Myung~Jin Choi, Antonio Torralba, and Alan~S Willsky.
\newblock A tree-based context model for object recognition.
\newblock {\em IEEE TPAMI}, 2012.

\bibitem[\protect\citeauthoryear{Cinbis and
  Sclaroff}{2012}]{cinbis2012contextual}
Ramazan Cinbis and Stan Sclaroff.
\newblock Contextual object detection using set-based classification.
\newblock {\em ECCV}, 2012.

\bibitem[\protect\citeauthoryear{Deng \bgroup \em et al.\egroup
  }{2009}]{deng2009imagenet}
Jia Deng, Wei Dong, Richard Socher, Li-Jia Li, Kai Li, and Li~Fei-Fei.
\newblock Imagenet: A large-scale hierarchical image database.
\newblock In {\em CVPR}. IEEE, 2009.

\bibitem[\protect\citeauthoryear{Ding \bgroup \em et al.\egroup
  }{2006}]{ding2006orthogonal}
Chris Ding, Tao Li, Wei Peng, and Haesun Park.
\newblock Orthogonal nonnegative matrix t-factorizations for clustering.
\newblock In {\em SIGKDD}, 2006.

\bibitem[\protect\citeauthoryear{Doersch \bgroup \em et al.\egroup
  }{2013}]{doersch2013mid}
Carl Doersch, Abhinav Gupta, and Alexei~A Efros.
\newblock Mid-level visual element discovery as discriminative mode seeking.
\newblock In {\em NIPS}, 2013.

\bibitem[\protect\citeauthoryear{Fan \bgroup \em et al.\egroup
  }{2008}]{fan2008integrating}
Jianping Fan, Yuli Gao, and Hangzai Luo.
\newblock Integrating concept ontology and multitask learning to achieve more
  effective classifier training for multilevel image annotation.
\newblock {\em IEEE TIP}, 2008.

\bibitem[\protect\citeauthoryear{Farhadi \bgroup \em et al.\egroup
  }{2009}]{farhadi2009describing}
Ali Farhadi, Ian Endres, Derek Hoiem, and David Forsyth.
\newblock Describing objects by their attributes.
\newblock In {\em CVPR}, 2009.

\bibitem[\protect\citeauthoryear{Feng and Bhanu}{}]{fengsemantic}
Linan Feng and Bir Bhanu.
\newblock Semantic concept co-occurrence patterns for image annotation and
  retrieval.
\newblock {\em IEEE TPAMI}.

\bibitem[\protect\citeauthoryear{He \bgroup \em et al.\egroup
  }{2016}]{he2016deep}
Kaiming He, Xiangyu Zhang, Shaoqing Ren, and Jian Sun.
\newblock Deep residual learning for image recognition.
\newblock In {\em CVPR}, 2016.

\bibitem[\protect\citeauthoryear{Johnson \bgroup \em et al.\egroup
  }{2016}]{johnson2016clevr}
Justin Johnson, Bharath Hariharan, Laurens van~der Maaten, Li~Fei-Fei,
  C~Lawrence Zitnick, and Ross Girshick.
\newblock Clevr: A diagnostic dataset for compositional language and elementary
  visual reasoning.
\newblock {\em arXiv}, 2016.

\bibitem[\protect\citeauthoryear{Juneja \bgroup \em et al.\egroup
  }{2013}]{juneja2013blocks}
Mayank Juneja, Andrea Vedaldi, CV~Jawahar, and Andrew Zisserman.
\newblock Blocks that shout: Distinctive parts for scene classification.
\newblock In {\em CVPR}, 2013.

\bibitem[\protect\citeauthoryear{Krizhevsky \bgroup \em et al.\egroup
  }{2012}]{krizhevsky2012imagenet}
Alex Krizhevsky, Ilya Sutskever, and Geoffrey~E Hinton.
\newblock Imagenet classification with deep convolutional neural networks.
\newblock In {\em NIPS}, 2012.

\bibitem[\protect\citeauthoryear{Lan \bgroup \em et al.\egroup
  }{2013}]{lan2013subcategories}
Tian Lan, Michalis Raptis, Leonid Sigal, and Greg Mori.
\newblock From subcategories to visual composites.
\newblock In {\em ICCV}, 2013.

\bibitem[\protect\citeauthoryear{Li \bgroup \em et al.\egroup
  }{2010}]{li2010object}
Li-Jia Li, Hao Su, Li~Fei-Fei, and Eric~P Xing.
\newblock Object bank: A high-level image representation for scene
  classification.
\newblock In {\em NIPS}, 2010.

\bibitem[\protect\citeauthoryear{Li \bgroup \em et al.\egroup
  }{2012}]{li2012automatic}
Congcong Li, Devi Parikh, and Tsuhan Chen.
\newblock Automatic discovery of groups of objects for scene understanding.
\newblock In {\em CVPR}, 2012.

\bibitem[\protect\citeauthoryear{Miettinen \bgroup \em et al.\egroup
  }{2008}]{miettinen2008discrete}
Pauli Miettinen, Taneli Mielik{\"a}inen, Aristides Gionis, Gautam Das, and
  Heikki Mannila.
\newblock The discrete basis problem.
\newblock {\em IEEE TKDE}, 2008.

\bibitem[\protect\citeauthoryear{Paatero and
  Tapper}{1994}]{paatero1994positive}
Pentti Paatero and Unto Tapper.
\newblock Positive matrix factorization: A non-negative factor model with
  optimal utilization of error estimates of data values.
\newblock {\em Environmetrics}, 1994.

\bibitem[\protect\citeauthoryear{Pandey and Lazebnik}{2011}]{pandey2011scene}
Megha Pandey and Svetlana Lazebnik.
\newblock Scene recognition and weakly supervised object localization with
  deformable part-based models.
\newblock In {\em ICCV}, 2011.

\bibitem[\protect\citeauthoryear{Parizi \bgroup \em et al.\egroup
  }{2012}]{parizi2012reconfigurable}
Sobhan~Naderi Parizi, John~G Oberlin, and Pedro~F Felzenszwalb.
\newblock Reconfigurable models for scene recognition.
\newblock In {\em CVPR}, 2012.

\bibitem[\protect\citeauthoryear{Patterson and Hays}{2012}]{patterson2012sun}
Genevieve Patterson and James Hays.
\newblock Sun attribute database: Discovering, annotating, and recognizing
  scene attributes.
\newblock In {\em CVPR}, 2012.

\bibitem[\protect\citeauthoryear{Quattoni and
  Torralba}{2009}]{quattoni2009recognizing}
Ariadna Quattoni and Antonio Torralba.
\newblock Recognizing indoor scenes.
\newblock In {\em CVPR}, 2009.

\bibitem[\protect\citeauthoryear{Rabinovich \bgroup \em et al.\egroup
  }{2007}]{rabinovich2007objects}
Andrew Rabinovich, Andrea Vedaldi, Carolina Galleguillos, Eric Wiewiora, and
  Serge Belongie.
\newblock Objects in context.
\newblock In {\em ICCV}, 2007.

\bibitem[\protect\citeauthoryear{Redmon and Farhadi}{2017}]{redmon2016yolo9000}
Joseph Redmon and Ali Farhadi.
\newblock Yolo9000: better, faster, stronger.
\newblock {\em CVPR}, 2017.

\bibitem[\protect\citeauthoryear{Simonyan and
  Zisserman}{2014}]{simonyan2014very}
Karen Simonyan and Andrew Zisserman.
\newblock Very deep convolutional networks for large-scale image recognition.
\newblock {\em arXiv}, 2014.

\bibitem[\protect\citeauthoryear{Singh \bgroup \em et al.\egroup
  }{2012}]{singh2012unsupervised}
Saurabh Singh, Abhinav Gupta, and Alexei~A Efros.
\newblock Unsupervised discovery of mid-level discriminative patches.
\newblock In {\em ECCV}. 2012.

\bibitem[\protect\citeauthoryear{Song \bgroup \em et al.\egroup
  }{2015}]{song2015sun}
Shuran Song, Samuel~P Lichtenberg, and Jianxiong Xiao.
\newblock Sun rgb-d: A rgb-d scene understanding benchmark suite.
\newblock In {\em CVPR}, 2015.

\bibitem[\protect\citeauthoryear{Sun and Ponce}{2013}]{sun2013learning}
Jian Sun and Jean Ponce.
\newblock Learning discriminative part detectors for image classification and
  cosegmentation.
\newblock In {\em ICCV}, 2013.

\bibitem[\protect\citeauthoryear{Szegedy \bgroup \em et al.\egroup
  }{2015}]{szegedy2015going}
Christian Szegedy, Wei Liu, Yangqing Jia, Pierre Sermanet, Scott Reed, Dragomir
  Anguelov, Dumitru Erhan, Vincent Vanhoucke, and Andrew Rabinovich.
\newblock Going deeper with convolutions.
\newblock In {\em CVPR}, 2015.

\bibitem[\protect\citeauthoryear{Torresani \bgroup \em et al.\egroup
  }{2010}]{torresani2010efficient}
Lorenzo Torresani, Martin Szummer, and Andrew Fitzgibbon.
\newblock Efficient object category recognition using classemes.
\newblock {\em ECCV}, 2010.

\bibitem[\protect\citeauthoryear{Zhang \bgroup \em et al.\egroup
  }{2007}]{zhang2007binary}
Zhongyuan Zhang, Tao Li, Chris Ding, and Xiangsun Zhang.
\newblock Binary matrix factorization with applications.
\newblock In {\em ICDM}, 2007.

\bibitem[\protect\citeauthoryear{Zhou \bgroup \em et al.\egroup
  }{2016}]{zhou2016learning}
Bolei Zhou, Aditya Khosla, Agata Lapedriza, Aude Oliva, and Antonio Torralba.
\newblock Learning deep features for discriminative localization.
\newblock In {\em CVPR}, 2016.

\bibitem[\protect\citeauthoryear{Zhou \bgroup \em et al.\egroup
  }{2017a}]{zhou2017places}
Bolei Zhou, Agata Lapedriza, Aditya Khosla, Aude Oliva, and Antonio Torralba.
\newblock Places: A 10 million image database for scene recognition.
\newblock {\em IEEE TPAMI}, 2017.

\bibitem[\protect\citeauthoryear{Zhou \bgroup \em et al.\egroup
  }{2017b}]{zhou2017scene}
Bolei Zhou, Hang Zhao, Xavier Puig, Sanja Fidler, Adela Barriuso, and Antonio
  Torralba.
\newblock Scene parsing through ade20k dataset.
\newblock In {\em CVPR}, 2017.

\end{thebibliography}
}
\end{document}